\def\year{2021}\relax
\documentclass[letterpaper]{article} 
\usepackage{aaai21}  
\usepackage{times}  
\usepackage{helvet} 
\usepackage{courier}  
\usepackage[hyphens]{url}  
\usepackage{graphicx} 
\usepackage[switch]{lineno}
\usepackage{natbib}  
\usepackage{caption} 
\usepackage{graphicx}
\usepackage{amsmath}
\usepackage{amsthm}
\usepackage{subcaption}
\usepackage{amssymb}
\usepackage{multirow}
\usepackage{booktabs}
\newcommand{\R}{\mathbb{R}}

\makeatletter
\newcommand{\ssymbol}[1]{$^{\@fnsymbol{#1}}$}
\makeatother

\usepackage{bbm}
\usepackage{color}

\def \1{{\textbf 1}}

\def \y{{\textbf y}}

\def \W{{\textbf W}}

\def \A{{\textbf A}}

\usepackage{hyperref}

\title{Audio-Oriented Multimodal Machine Comprehension: Task, Dataset and Model}
\author{
    Zhiqi Huang\textsuperscript{\rm 1\thanks{Equal Contributions.}},
    Fenglin Liu\textsuperscript{\rm 1\footnotemark[1]},
    Xian Wu\textsuperscript{\rm 2},
    Shen Ge\textsuperscript{\rm 2},
    Helin Wang\textsuperscript{\rm 1},
    Wei Fan\textsuperscript{\rm 2},
    Yuexian Zou\textsuperscript{\rm 1,3\thanks{Corresponding Author.}}
    \\
}
\affiliations{
    \textsuperscript{\rm 1} ADSPLAB, School of ECE, Peking University, China \\
    \textsuperscript{\rm 2} Tencent, China \\ 
    \textsuperscript{\rm 3} Peng Cheng Laboratory, China \\
    \{zhiqihuang, fenglinliu98, wanghl15, zouyx\}@pku.edu.cn, \{kevinxwu, shenge, davidwfan\}@tencent.com \\   
}

\begin{document}

\maketitle

\begin{abstract}
While Machine Comprehension (MC) has attracted extensive research interests in recent years, existing approaches mainly belong to the category of Machine Reading Comprehension task which mines textual inputs (paragraphs and questions) to predict the answers (choices or text spans). However, there are a lot of MC tasks that accept audio input in addition to the textual input, e.g. English listening comprehension test. In this paper, we target the problem of Audio-Oriented Multimodal Machine Comprehension, and its goal is to answer questions based on the given audio and textual information. To solve this problem, we propose a Dynamic Inter- and Intra-modality Attention (DIIA) model to effectively fuse the two modalities (audio and textual). DIIA can work as an independent component and thus be easily integrated into existing MC models. Moreover, we further develop a Multimodal Knowledge Distillation (MKD) module to enable our multimodal MC model to accurately predict the answers based only on either the text or the audio. As a result, the proposed approach can handle various tasks including: Audio-Oriented Multimodal Machine Comprehension, Machine Reading Comprehension and Machine Listening Comprehension, in a single model, making fair comparisons possible between our model and the existing unimodal MC models. Experimental results and analysis prove the effectiveness of the proposed approaches. First, the proposed DIIA boosts the baseline models by up to 21.08\% in terms of accuracy; Second, under the unimodal scenarios, the MKD module allows our multimodal MC model to significantly outperform the unimodal models by up to 18.87\%, which are trained and tested with only audio or textual data.
\end{abstract}
\vspace{-0.1in}

\section{Introduction}
\begin{figure*}[t]
	\centering  
	\includegraphics[width=1\textwidth]{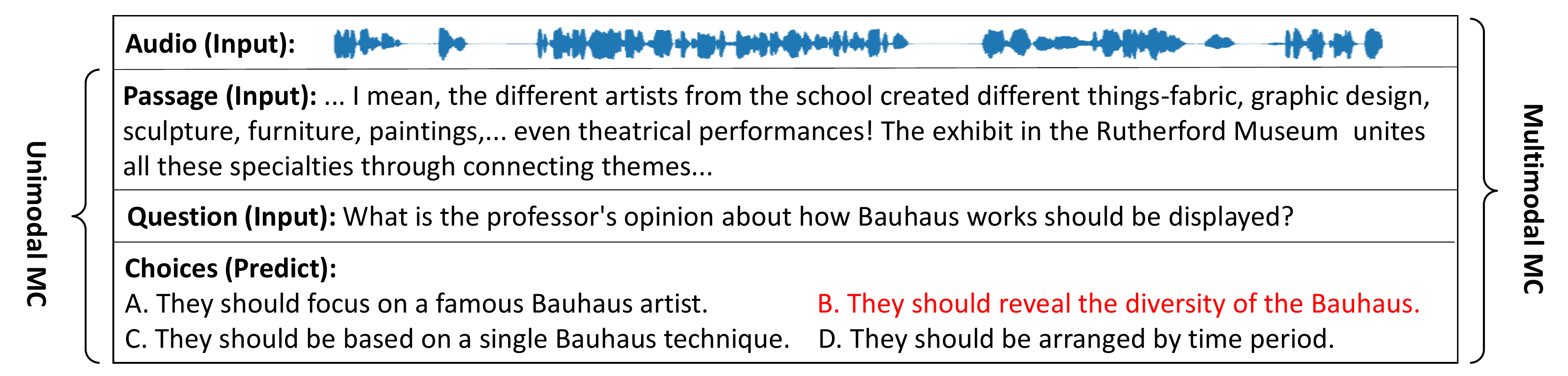}
	\caption{Comparison between the conventional unimodal machine comprehension and the proposed audio-oriented multimodal machine comprehension. The red colored text indicate the ground-truth answer.}
	\label{fig:example}
\end{figure*}

Recently, there is a surge of research interests in Machine Comprehension (MC), which aims to teach the machine to answer questions after giving comprehension materials~\cite{Nguyen2016MARCO,Rajpurkar2016SQuAD,lai2017large}. As shown in Figure~\ref{fig:example}, conventional MC system accepts unimodal textual inputs, and predict the corresponding answers to the given multiple-choice questions. By adopting various deep learning techniques, many models have been developed for the MC problem and are proven to be effective~\cite{Liu2019survey,Qiu2019survey2}.

However, conventional MC only focus on accepting single modal textual inputs, while in real life many multimodal (audio and textual modalities) scenarios exist, such as playing music with lyrics, and taking a listening examination in TOEFL, etc.
Moreover, multimodal inputs often convey more information than single modality inputs, and it is easy to make wrong judgments under single modal scenarios.
For example, people could express opposite intentions by using different tones to say the sentence ``that's interesting''. 
If the emphasis is put on the word ``interesting!'', he/she may really be interested and wants to know more; on the other hand, if the whole sentence is expressed intermittently, e.g., ``that's... um... interesting.'', he/she may not be interested at all.
Another example in English exams like TOEFL, i.e., ”He hasn’t seen his parent four years!” and ”He hasn’t seen his parent for years!”, may only be distinguished by their different sound emphasis, showing a clear trap in the audio.
As a result, different tones of voice could lead to different meanings, even when the textual sentences are nearly identical.
Inspired by these real-world applications and observations,  we propose the novel problem of \textit{Audio-Oriented Multimodal Machine Comprehension}. As shown in Figure~\ref{fig:example}, the novel problem requires the system to consider both the audio and textual modality inputs in selecting the correct answers. 
Compared to the conventional setting of unimodal based MC, our new problem poses two fundamental challenges.
First, in the learning of multimodal tasks, due to the great disparities between the textual and the audio domains, plus the distinct features of the modalities, one core challenge is to effectively bridge the gap between textual and the audio domains and learn an effective fusion of multimodality features \cite{lu2016coattention,Gao2019DFAF,liu2019mia,liu2020bridging}.
Second, due to the abundance of unimodal (textual or audio) real-world application scenarios, e.g., the conventional machine reading comprehension (textual) \cite{lai2017large} and the end-to-end spoken language understanding (audio) \cite{Dmitriy2018Towards}\nocite{huang2021SLU, huang2020SLU, Zhou2020SLU}, it is necessary to enable the multimodal MC model to work in the unimodal scenarios. In other words, we should empower the proposed multimodal MC model with the capability to accurately predict the answers based only on the textual or only on audio input.

To tackle the first challenge, we propose a novel Dynamic Inter- and Intra-modality Attention (DIIA) model to better capture the high-level interactions between audio and textual features, resulting in an efficient multimodality feature fusion to answer the questions accurately.
As shown in Figure~\ref{model}, the DIIA integrates the self-attention and co-attention to learn the inter- and intra-relationships between audio and textual modalities in an effective manner \cite{liu2019mia,liu2019glied,liu2020aimNet}.
The core intuition behind our motivation is that, each textual word should obtain information not only from its associated audio information but also from related words/phrases to infer the answer to the question, and so do audio.
To tackle the second challenge, based on our proposed DIIA model, we further develop a Multimodal Knowledge Distillation (MKD) module, which transfers representative knowledge from multimodal to either textual or audio modality.
In implementation, as shown in Figure~\ref{model}, the learned multimodal representations from our pre-trained multimodal MC model, i.e., DIIA, are used to guide the learning of unimodal representations.
As a result, the MKD module associates the multimodal knowledge hidden behind the fused multimodal features to facilitate the understanding of unimodal information.
The design of the proposed MKD allows our approach to be applied to scenarios where only single modal data is available. In other words, our approach can accurately answer the questions based only on input audio or input text, so that our approach can be used for fair comparisons with existing textual based MC models.

Moreover, to better handle this problem, we also collect two audio-oriented multimodal machine comprehension datasets, i.e., L-TOEFL and CET, from English listening tests, which contain questions and answers in the form of text, as well as the comprehension passages in both textual and audio modalities.
The extensive experiments and analysis on the proposed L-TOEFL and CET datasets validate our arguments and prove the effectiveness of our approach. 

Overall, our main contributions are as follows: 
\begin{itemize}
	\item We propose the audio-oriented multimodal machine comprehension task, which requires the system to understand both input audio and textual information together, rather than only use textual information in previous works. We also assemble two audio-oriented multimodal machine comprehension datasets (L-TOEFL and CET) for the task. 
	
	\item A novel Dynamic Inter- and Intra-modality Attention (DIIA) model is proposed to obtain multimodality fusion by interleaving inter- and intra-modality feature-level fusion. Such a framework provides a solid bias for the audio-oriented multimodal machine comprehension task.
	
	\item We further develop the Multimodal Knowledge Distillation (MKD) module, which can transfer representative knowledge from multimodal to either textual modal or audio modal, enabling a single multimodal MC model to accurately predict the answers based only on the text or audio. As a result, our model can handle various tasks at the same time with a single model.
	
	\item The extensive experiments show consistent performance gains achieved by the proposed novel DIIA over baseline systems. The experiments also show that the proposed MKD module enables the multimodal MC model to be applied in the unimodal scenarios and outperform the conventional unimodal models significantly.
\end{itemize}

	\section{Related Work}
	The related works are introduced from two aspects: 1) Multi-Choices Machine Comprehension and 2) Machine Comprehension of Spoken Content.
	
	\subsection{Multi-Choices Machine Comprehension}
	There are four types of Machine Comprehension (MC)~\cite{chen2018phd}, including cloze-style~\cite{Hermann2015Mail}, multi-choices~\cite{lai2017large}, span extraction~\cite{Rajpurkar2016SQuAD,Rajpurkar2018SQuAD2}, and generative format~\cite{Kocisk2018Narrativeqa}. In this paper, we focus on multi-choices machine comprehension: the goal is to find the only correct answer in the multiple (usually 4) choices based on the given inputs, i.e., passage, question, and multiple choices.

	\subsection{Machine Comprehension of Spoken Content}
	\citet{tseng2016toefl} proposed to deal with the MC task of spoken content by first employ an ASR model~\cite{Yu2017ASR} to recognize speech into text, then, an MC model is designed to process the ASR transcriptions for selecting the correct answer out of 4 choices.
    Although such method take the spoken content into consideration, the system is still a text-based MC system. 
    \citet{Chuang2020SpeechBERT,Kuo2020SpokenMCQA}
	studied the end-to-end spoken question answering problem by introducing a pre-trained modality fusion model learned from audio and text, however, in the inference phase, they need input with both modalities while ours can predict with only audio or only text input.
    And there are also some related work studied different tasks~\cite{Yang2003VQA} or introduced auxiliary tool~\cite{Zhang2018KGQA}.  
	
	\section{Approach}
\begin{figure*}[t]
		\centering
		\includegraphics[width=1\textwidth]{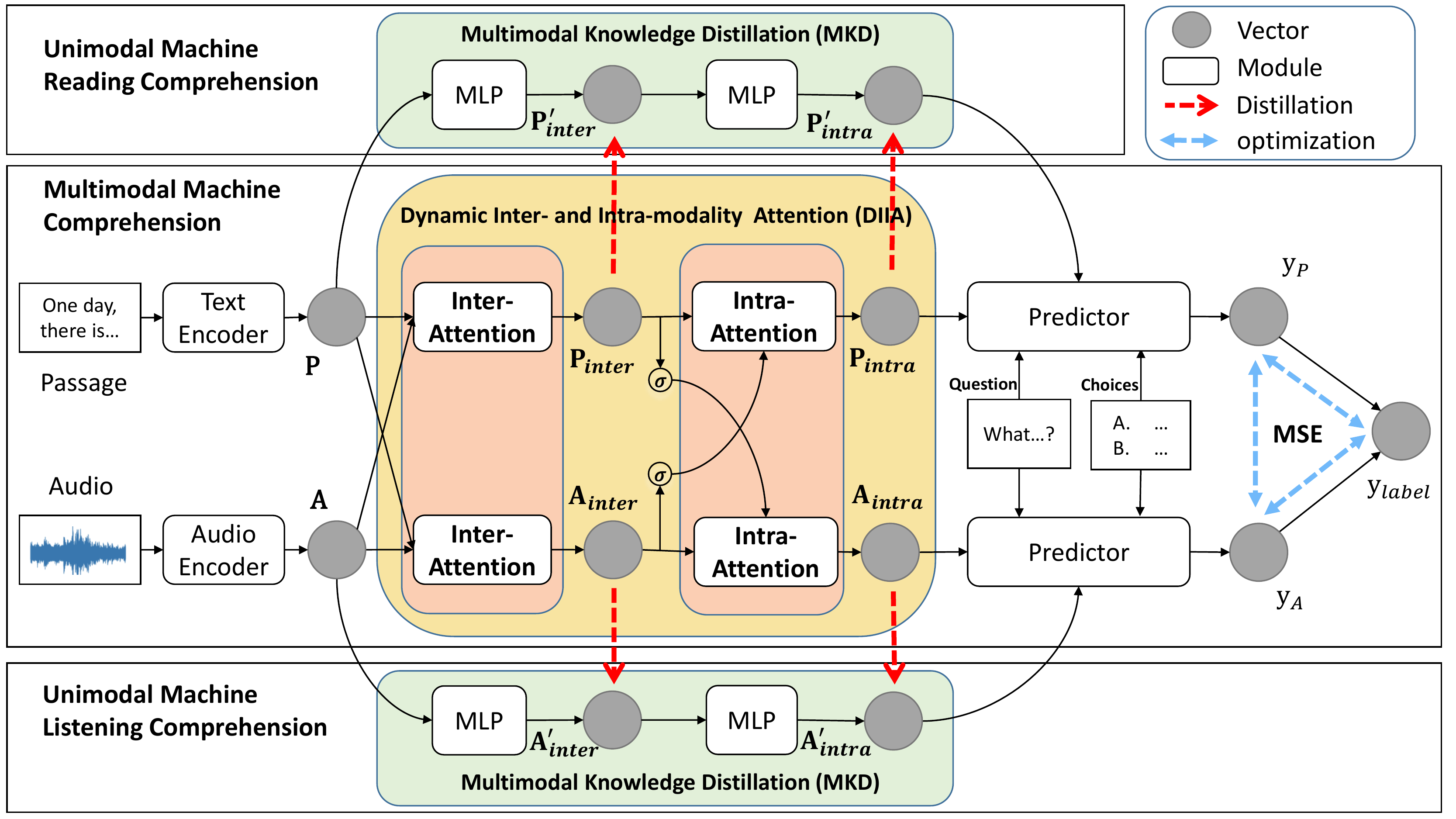}\\
		\caption{The architecture of our proposed approach. 
			The DIIA model consists of the inter-modality attention module and the intra-modality attention module, aiming to capture the correlation and build the relationship between audio and textual modalities.
			The blue dashed lines represent the MSE loss between $\y_{A}$, $\y_{P}$ (the training label) and $\y_{label}$ (the ground truth label).
			The red dashed lines represent the hidden states based distillation. The predictor is adapted from the existing machine comprehension models, such as Co-Matching \cite{Wang2018comatch}.}
		\vspace{-0.11in}
        \label{model}
	\end{figure*}

	In this section, we first formulate the conventional machine comprehension (MC) and the proposed multimodal machine comprehension problems; then we describe the proposed approach in detail.
	Specifically, to better capture the high-level interactions between audio and textual features, and generate efficient multimodality feature fusion to accurately answer questions, we propose the novel Dynamic Inter- and Intra-modality Attention (DIIA) model.
	Based on our DIIA model, we further propose the Multimodal Knowledge Distillation (MKD) to enable our multimodal MC model to accurately predict the answers based only on the text or audio. Figure~\ref{model} illustrates the proposed approach.

	\subsection{Problem Formulation}
	In this section, we formulate the problems of conventional multi-choices machine comprehension and the proposed multi-choices multimodal machine comprehension.
	
	\subsubsection{Problem Formulation of Conventional Multi-Choices Machine Comprehension Task}
	
	Taking a passage $\textbf{P}_\text{I}$ as input, the goal of conventional multi-choices machine comprehension is to predict the correct choice $\textbf{C}_\text{ans}$ based on the given questions $\textbf{Q}$ and candidate choices $\textbf{C}_\text{candidate}$.
	Some well-performing frameworks~\cite{Wang2018comatch,Dhingra2017GAReader,Devlin2019BERT} normally include a text encoder and an answer predictor, which can be formulated as:
	\begin{linenomath*}
	\begin{equation*}
	\begin{aligned} 
	\footnotesize
	\text{Text Encoder}  &: \textbf{P}_\text{I} \rightarrow \textbf{P} \\
	\text{Answer Predictor} &: {\textbf{P}, \textbf{Q}, \textbf{C}_\text{candidate}} \rightarrow \textbf{C}_\text{ans}
	\end{aligned}
	\end{equation*}
	\end{linenomath*}
	The text encoder aims to generate the textual features $\textbf{P}$  of the input passage $\textbf{P}_\text{I}$. In implementation, given an input sequence with length $N$, the text features are usually generated by 300-dimensional word embeddings with GloVe vectors~\cite{Pennington2014Glove}, and represented as $\textbf{P} \in \R^{N \times d}$ ($d = 300$). 
	The answer predictor, e.g., co-matching \cite{Wang2018comatch}, is used to predict the correct choice $\textbf{C}_\text{ans}$ from the given questions $\textbf{Q}$, the candidate choices $\textbf{C}_\text{candidate}$ and $\textbf{P}$. Given the ground truth choice, we can simply train the framework by minimizing training loss, e.g., cross-entropy loss.
	
	\subsubsection{Problem Formulation of Multimodal Multi-Choices Machine Comprehension Task}
	The main difference between the audio-oriented multimodal machine comprehension and conventional machine comprehension is the different available information that can be used to predict the correct answer.
	Specifically, for audio-oriented multimodal machine comprehension task, it requires the system to further consider the audio information $\textbf{A}_\text{I}$ when selecting the correct answer, which can be formulated as: 
	\begin{linenomath*}
	\begin{equation*}
	\begin{aligned} 
	\footnotesize
	\text{Text Encoder}  &: \textbf{P}_\text{I} \rightarrow \textbf{P} \\
	\text{Audio Encoder}  &: \textbf{A}_\text{I} \rightarrow \textbf{A} \\
	\text{Answer Predictor} &: {\textbf{P}, \textbf{A}, \textbf{Q}, \textbf{C}_\text{candidate}} \rightarrow \textbf{C}_\text{ans}
	\end{aligned}
	\end{equation*}
	\end{linenomath*}
	where $\textbf{A}$ denotes the extracted audio features from the input audio $\textbf{A}_\text{I}$. In implementation, we adopt VGGish~\cite{Hershey2017vggish} pre-trained on AudioSet~\cite{Gemmeke2017AudioSet} to extract the audio features, represented as $\textbf{A} \in \R^{M \times d}$.

	\subsection{Dynamic Inter- and Intra-modality Attention}

	Our Audio-Oriented Multimodal Machine Comprehension task requires the MC system to understand both input audio and textual information, thus learning fine-grained joint representations of audio and text are of paramount importance. In other words, it is vital to learn the alignments and relationships between audio and textual modalities.
	To this end, as shown in Figure~\ref{model}, we propose the Dynamic Inter- and Intra-modality Attention (DIIA) to effectively fuse the audio and textual features before predicting answers.
	In particular, inspired by the success of Multi-Head Attention (MHA)~\cite{Vaswani2017Attention}, we refer to the MHA mechanism and propose the DIIA model, which consists of an Inter-modality Attention module and an Intra-modality Attention module, to learn the inter- and intra-relationships of audio and textual modalities in an effective manner.

	\subsubsection{Multi-Head Attention} 
	In order to extract the relationship between the intra-modality and inter-modality of audio features and textual features, we adopt the Multi-Head Attention (MHA) \cite{Vaswani2017Attention}, which compute the association weights between different features. The attention mechanism allows probabilistic many-to-many relations instead of monotonic relations, as in \citet{xu2015show,liu2019mia}.
	The following MHA consists of $n$ parallel heads and each head is represented as scaled dot-product attention. 
    \begin{linenomath*}
	\begin{equation*}
	\begin{aligned} 
    \text{Att}(\textbf{Q,K,V}) = \text{softmax}\left(\frac{\textbf{Q}\text{W}_\text{Q}(\textbf{K}\text{W}_\text{K})^{T}}{\sqrt{d_k}}\right)\textbf{V}\text{W}_\text{V}
	\end{aligned}
	\end{equation*}
	\end{linenomath*}   
    where $\textbf{Q} \in \mathbb{R}^{l \times d}$, $\textbf{K} \in \mathbb{R}^{{k} \times d}$ and $\textbf{V} \in \mathbb{R}^{{k} \times d}$ represent respectively the query matrix, the key matrix and the value matrix; The $l$ and $k$ denote the length of the query and key/value, respectively; $\text{W}_\text{Q}, \text{W}_\text{K}, \text{W}_\text{V} \in \mathbb{R}^{d \times d_k}$ are the learnable parameters of linear transformations and ${d}_{k} = d / {n} $ is the scaling factor, where $n$ is the number of heads.
    
    Following the multi-head attention is a fully-connected Feed-Forward Network (FFN), which is defined as follows:
	\begin{linenomath*}
	\begin{equation*}
	\begin{aligned} 
    \text{FFN}(\textbf{x}) = \text{ReLU}(\textbf{x}\text{W}_\text{f}+{b}_\text{f})\text{W}_\text{ff}+{b}_\text{ff} 
	\end{aligned}
	\end{equation*}
	\end{linenomath*}
    where 
    $\text{W}_\text{f}$ and $\text{W}_\text{ff}$ denote matrices for linear transformation; $b_\text{f}$ and ${b}_\text{ff}$ represent the bias terms. Each sub-layer, i.e., MHA and FFN, is followed by an operation sequence\footnote{For conciseness, the operation sequence in this paper is omitted.} of dropout \cite{srivastava2014dropout}, shortcut connection \cite{he2016deep}, and layer normalization \cite{ba2016layernormalization}.
    
	We take advantage of the MHA to implement the idea of learning the inter- and intra-relationships of audio and textual modalities.
	
	\subsubsection{Inter-modality Attention} 
	1) To represent textual features $\textbf{P}$ with high quality, we need to find the most relevant audio descriptions $\textbf{A}$ to identify the direct relations between audio and text.
	2) Similarly, we need to find the most relevant textual features $\textbf{P}$ to summarize the properties of the audio features $\textbf{A}$.
	
	According to the attention theorem  \cite{Vaswani2017Attention}, taking the first situation as example, the textual features $\textbf{P} \in \mathbb{R}^{N \times d}$ serve as query, and the audio features $\textbf{A} \in \mathbb{R}^{M \times d}$ server as key and value. Consequently, the result $\mathbf{A_\text{inter}} \in \mathbb{R}^{M \times d}$ turns out to be a set of attended audio features for textual features:
	\begin{linenomath*}
	\begin{equation*}
	\begin{aligned} 
    \mathbf{A_\text{inter}} = 
    \text{FFN}(\text{MHA}(\textbf{A},\textbf{P},\textbf{P})) 
    \end{aligned}
	\end{equation*}
	\end{linenomath*}

	Similarly, the $\mathbf{P_\text{inter}} \in \mathbb{R}^{N \times d}$ can be computed as follow:
	\begin{linenomath*}
	\begin{equation*}
	\begin{aligned} 
    \mathbf{P_\text{inter}} = 
    \text{FFN}(\text{MHA}(\textbf{P},\textbf{A},\textbf{A}))
	\end{aligned}
	\end{equation*}
	\end{linenomath*}
	Now we can assume that the relation between the audio and textual features are built and represent the two updated features as $\mathbf{A}_\text{inter}  \in \mathbb{R}^{M \times d}$ and $\mathbf{P}_\text{inter}  \in \mathbb{R}^{N \times d}$.

	\paragraph{Intra-modality Attention} 
	After the inter-modality attention module, the cross-modal relations between audio and text have been modeled. 
	However, information from different modalities may have varying predictive power and noise.
	We argue that modeling relationships in a single modality can make up for this deficiency.
	The intra-modality attention module explores how the knowledge learned from two modalities can be fused in an appropriate way to help the training of the multimodal MC model.
    We adapt the following formula to learn the intra-relationships of audio and textual features:
	\begin{linenomath*}
	\begin{equation*}
	\begin{aligned} 
    \mathbf{A_\text{intra}} &= 
    \text{FFN}(\text{MHA}(\mathbf{\hat{A}_\text{inter}},\mathbf{\hat{A}_\text{inter}}, \mathbf{A_\text{inter}})) \\
    \mathbf{P_\text{intra}} &= 
    \text{FFN}(\text{MHA}(\mathbf{\hat{P}_\text{inter}}, \mathbf{\hat{P}_\text{inter}}, \mathbf{P_\text{inter}}))
	\end{aligned}
	\end{equation*}
	\end{linenomath*}    
    It is worth noticing that to better promote the learning of intra-relationships \cite{hu2020senet,liu2018simnet}, we further design a conditional gate operation $\mathbf{G}$ to update the queries and keys. The process is defined as follows: 
	\begin{linenomath*}
	\begin{equation*}
	\begin{aligned}
    \mathbf{\hat{A}_\text{inter}} &= \left(1+\mathbf{G_{P}}\right) \odot \mathbf{A_\text{inter}} \\
    \mathbf{\hat{P}_\text{inter}} &= \left(1+\mathbf{G_{A}}\right) \odot \mathbf{P_\text{inter}}
    \end{aligned}
	\end{equation*}
	\end{linenomath*}
    where $\odot$ represents the element-wise multiplication. The conditional gate operation $\mathbf{G_{P}}$ and $\mathbf{G_{A}}$ are defined as:
	\begin{linenomath*}
	\begin{equation*}
	\begin{aligned} 
	\mathbf{G_{A}}&=\sigma\left(\operatorname{Avg\_pool}(\mathbf{A})\mathbf{W_A}\right) \\
	\mathbf{G_{P}}&=\sigma\left(\operatorname{Avg\_pool}(\mathbf{P})\mathbf{W_P}\right)
	\end{aligned}
	\end{equation*}
	\end{linenomath*}
    where the $\sigma$ and $\operatorname{Avg\_pool}$ denote the sigmoid function and average pooling, respectively.

	Through the formula, in the audio domain, the intra-modality attention learns salient audio groupings and integrates naturally related audio information. In the textual domain, it learns text collocations and has the ability to consider associations and collocations of sentences in the passage during answer predicting. The learned intra-relationships of audio and textual features are super beneficial for multimodal machine comprehension task.

	\subsubsection{Multimodal Knowledge Distillation} 
	The intuition of enabling a model to accurately predict the answer based only on either the text or the audio is that conventionally the model is only allowed to accept a single modality as input.
	However, due to the fact that the proposed DIIA have learned the fine-grained multimodal representations, inspired by the knowledge distillation technique~\cite{Romero2015distill}, we further introduce the Multimodal Knowledge Distillation Module (MKD) to distill the representative knowledge from multimodal learnt by DIIA to either textual modal or audio modal to enhance the input features with single modality only.
	In particular, the MKD consists of two Multi-Layer Perceptrons (MLPs). We represent the output of the MLPs as $\mathbf{A^{'}_\text{inter}}$, $\mathbf{A^{'}_\text{intra}}$ for audio distillation block and $\mathbf{P^{'}_\text{inter}}$, $\mathbf{P^{'}_\text{intra}}$ for passage distillation block, which are defined as:
	\begin{linenomath*}
    \begin{equation*}
    \begin{aligned}
    \mathbf{A^{'}_\text{inter}} &= 
    \text{MLP}(\mathbf{A}) ; \quad \mathbf{A^{'}_\text{intra}} = 
    \text{MLP}(\mathbf{A^{'}_\text{inter}}) \\
    \mathbf{P^{'}_\text{inter}} &= 
    \text{MLP}(\mathbf{P}) ; \quad \mathbf{P^{'}_\text{intra}} = 
    \text{MLP}(\mathbf{P^{'}_\text{inter}})
	\end{aligned}
    \end{equation*}
    \end{linenomath*}
	Then we apply the following formulas to distill the knowledge from the output of DIIA to the MKD, which can be computed with mean squared error (MSE) loss and be represented as:
	\begin{linenomath*}
	\begin{equation*}
	\begin{aligned} 
	\mathcal{L}_{\text{MKD}_A}&= \text{MSE} (\mathbf{A_\text{inter}}, \mathbf{A^{'}_\text{inter}})+\text{MSE} (\mathbf{A_\text{intra}}, \mathbf{A^{'}_\text{intra}}) \\ 
	\mathcal{L}_{\text{MKD}_P}&= \text{MSE} (\mathbf{P_\text{inter}}, \mathbf{P^{'}_\text{inter}})+\text{MSE} ( \mathbf{P_\text{intra}}, \mathbf{P^{'}_\text{intra}})
	\end{aligned}
	\label{eq:KD}
	\end{equation*}
	\end{linenomath*}
	
	In this way, with the help of MKD, we can use only one modal data input in the answer prediction process, while implicitly use the interaction information of the two modalities to enhance the unimodal representations.
	
	\begin{table*}[t]
		\begin{center}
		\small
			\begin{tabular}{@{}l|l|c c|c c|c c|c c|c c|c c@{}}
				\toprule
				& Datasets  & \multicolumn{6}{c|}{L-TOEFL} & \multicolumn{6}{c}{CET} \\ \cmidrule{2-14}
				Modalities & Methods
				& \multicolumn{2}{c|}{GAReader} & \multicolumn{2}{c|}{Co-Matching} & \multicolumn{2}{c|}{DCMN} 
				& \multicolumn{2}{c|}{GAReader} & \multicolumn{2}{c|}{Co-Matching} & \multicolumn{2}{c}{DCMN} \\
				\cmidrule{2-14}
		         & Data Partitions &  Dev & Test &  Dev & Test &  Dev  & Test&  Dev  & Test &  Dev  & Test &  Dev  & Test \\
				\midrule [\heavyrulewidth]
				\multirow{5}{*}{\rotatebox{90}{Unimodal}}
				\ \ \multirow{2}{*}{{Audio}}
				& Conventional & 31.29 & 30.63 & 34.55 & 33.01 & 43.42 & 41.61
				& 34.11 & 33.82 & 39.96 & 39.44 & 52.86 & 51.61 \\
				& \ w/ MKD & 43.76 & 43.31  & 48.34 & 48.09 & 62.40 & 60.51 
				& 47.11 & 46.44  & 52.66 & 52.50  & 65.39 & 63.99 \\ \cmidrule{2-14}
			    \ \ \ \ \ \ Text
				& Conventional & 36.35 & 36.19 & 38.26 & 38.75 & 46.85 & 46.77 
                & 41.73 & 41.22 & 47.73 & 47.90 & 60.07 & 58.68 \\
				& \ w/ MKD & 44.69 & 43.60  & 49.98 & 49.20  & 64.33 & 62.68 
				& 49.04 & 47.98  & 55.31 & 55.00  & 67.63 & 66.18 \\ 
				\midrule
				\multirow{4}{*}{Multimodal}
				& Shallow-fusion & 46.28 & 45.19  & 50.68 & 50.75 & 63.22 & 62.69
				& 51.33 & 48.62  & 57.81 & 56.11 & 66.52 & 64.39 \\
				& \ w/ Inter- & 47.15 & 46.11 & 52.40 & 51.34 & 64.01 & 63.77 & 53.13 & 50.74 & 58.03 & 56.50 & 68.94 & 67.80 \\
				& \ w/ Intra- & 45.98 & 45.60 & 51.93 & 51.14 & 63.79 & 62.92 & 52.05 & 50.22 & 57.85 & 56.49 & 67.35 & 66.01  \\
				& \ w/ DIIA & \textbf{48.36} & \textbf{47.78}  & \textbf{53.22} & \textbf{52.03}  & \textbf{65.94} & \textbf{63.68} & 
				\textbf{54.20} & \textbf{51.12}  & \textbf{59.01} & \textbf{57.83}  & \textbf{69.13} & \textbf{68.09} \\
				\bottomrule
			\end{tabular}  
		\end{center}
		\vspace{-0.05in}
		\caption{Performance (Accuracy (\%)) on the proposed L-TOEFL and CET datasets. Multimodal and Unimodal represent the input modalities we use for the models, i.e., audio and text, audio only, or text only.
		Conventional means the conventional models that only employ the predictor and the feature encoder.
		``w/ MKD'' means that we employ our proposed MKD module as well as our training method in the unimodal setting.
		Shallow fusion means directly add the two unimodal representations for prediction and bypass the DIIA module.
        }
        \vspace{-0.15in}
		\label{tab:result}
	\end{table*}
	
	\subsection{Implementation}
	From the above process, the proposed DIIA model focuses on learning the relationships between audio and textual features to obtain the multimodal representations, and the MKD module can distill the multimodal knowledge learned from the DIIA.
	In this section, we describe the training process detail of our approach by introducing three applied problems, i.e., Multimodal Machine Comprehension, Unimodal Machine Reading Comprehension and Unimodal Machine Listening Comprehension.
	
	\subsubsection{Multimodal Machine Comprehension}
	As shown in Figure~\ref{model}, 
	the outputs of the predictors of multimodal machine comprehension problem are $\y_{A}$ and $\y_{P}$ for the audio and textual features, respectively. Given the ground truth  $\y_{label}$, we adopt the Cross-Entropy (CE) loss function to optimize our multimodal MC model, including the DIIA. The optimization is defined as:
	\begin{linenomath*}
	\begin{equation*}
	\begin{aligned}
	\ell_{pred}(\y_{A},\y_{label})= \text{CE}(\y_{A},\y_{label})
	\\
	\ell_{pred}(\y_{P},\y_{label})= \text{CE}(\y_{P},\y_{label})
	\end{aligned}
	\end{equation*}
	\end{linenomath*}
	Specifically, for better optimization, we further distill the knowledge loss on the logits to narrow the distance between the audio logits and the textual logits through MSE loss:
	\begin{linenomath*}
	\begin{equation*}
	\begin{aligned} 
	\ell_{pred} ( \y_{A}, \y_{P}) = \text{MSE}(\y_{A},\y_{P})
	\end{aligned}
	\end{equation*}
	\end{linenomath*}
	
	Overall, the final objective loss function is computed as:
	\begin{linenomath*}
	\begin{equation*}
	\begin{aligned} 
	\mathcal{L}_{1}=& \ell_{pred} ( \y_{A}, \y_{label}) + \ell_{pred} ( \y_{P}, \y_{label}) + \ell_{pred}( \y_{A},\y_{P}) 
	\end{aligned}
	\end{equation*}
	\end{linenomath*}
	At the testing stage, the input $\textbf{P}$ and $\textbf{A}$ are sent to the DIIA to obtain the $\mathbf{P_\text{intra}}$ and $\mathbf{A_\text{intra}}$, then sent to the predictor to obtain the $\y_{A}$ and $\y_{P}$. Finally, we add the two predicted logits as the final predicted logits to predict the answer.

	\subsubsection{Unimodal Machine Reading Comprehension}
	For practical use, we further propose to enable our multimodal MC model to accurately predict the answers based only on the text, which means that we only use unimodal textual information in inference.
	Because the unimodal scenario requires no modal interaction, we remove the DIIA, instead, we introduce an MKD module to transfer the representative multimodal knowledge learned from our DIIA to the textual modality (see Figure~\ref{model}).
	Specifically, at the training stage, we first directly adopt the pre-trained multimodal MC model. Next, we freeze the parameters of the text encoder, DIIA and predictor, and use the proposed $\mathcal{L}_{\text{MKD}_P}$ to train the MKD. 
	At the testing stage, we obtain the input passage features of the predictor $\mathbf{\hat{P}_\text{inter}}$ through the MKD module, and then output the predicted logits $\y_{P}$.
	
	In this way, our model can work on unimodal input in practice 
	with multimodal information being introduced during the distillation training process.
	So that our model can be compared fairly with conventional MC models.
	
	\subsubsection{Unimodal Machine Listening Comprehension}
	Similar to the Unimodal Machine Reading Comprehension scenario, we 
	train the MKD using $\mathcal{L}_{\text{MKD}_A}$ with the pre-trained multimodal MC model froze. And generate the predicted logits $\y_{A}$ with the trained MKD at the test stage.

	\vspace{-0.09in}
	\section{Experiments}
	We describe the collected datasets and the training details, followed by the evaluation of the proposed approach.

	\subsection{Datasets}
	In this paper, to address the audio-oriented multimodal MC problem, we propose the L-TOEFL and CET datasets, where L-TOEFL is collected from the TOEFL Educational Testing Service (TOEFL ETS)\footnote{https://www.ets.org/toefl/}, which is an English ability test designed to measure the ability to listen for basic comprehension, pragmatic understanding and synthesizing information, and CET is collected from the College English Test (CET)\footnote{http://cet.neea.edu.cn/}, a national English as a foreign language test in the People's Republic of China.
	Designed by educational experts, L-TOEFL and CET datasets aim to test non-native English speakers using various types of complicated questions, and the passages are divided into types of conversation and lecture.
	 Specifically, L-TOEFL and CET are collected from a total of 106 official examinations, with each official examination contains 1 to 6 passage(s) with corresponding audio, each passage contains 1 to 6 question(s), and each question is accompanied with 4 choices. Thus, our datasets consist of 4-attributes pair: $\{$audio, passage, question, 4 answer choices with the correct one$\}$. After deleting some improper pairs, such as multiple answers (more than one correct answer), etc., we get a total of 2,200 such A-P-Q-C pairs.
	 We randomly divide the L-TOEFL and CET datasets into 1000/162/162 and 657/110/109 examples as for train/dev/test data partitioning, respectively, following ratios of 0.75/0.125/0.125. The amount of collected English listening test data set is larger than the one used in~\citet{tseng2016toefl} (963 examples).  
	
	\subsection{Experimental Settings}
	The encoder takes the audio features and text features as input.
	In implementation, we adopt the VGGish~\cite{Hershey2017vggish} pre-trained on AudioSet~\cite{Gemmeke2017AudioSet} to extract the audio features denoted as $\A \in \R^{M \times 300}$.
	First, we resample the raw audio file to the rate assumed by VGGish, then generate a 128-dimensional embedding of each AudioSet segment.
	After that, we employ a linear transformation to map the dimension from 128 to 300.
	Thus, the feature dimension of each audio is ${frame \times 300}$ where the ${frame}$ range from 109 to 469.
	Given an input sequence with length $N$, the text features are initialized by 300-dimensional word embeddings with GloVe vectors~\cite{Pennington2014Glove} denoted as $\W \in \R^{N \times 300}$.
	Considering the actual length of the datasets, we set the maximum length $M$ and $N$ as 384 empirically.
	We adopt the Adam optimizer for optimizing the parameters, with a mini-batch size of 12 and initial learning rate of 0.001. 
	After training 100 epochs, we select the model which works the best on the dev set, and then evaluate it on the test set in terms of accuracy (\%).

	\begin{figure*}[t]
		\centering
		\includegraphics[width=1\textwidth]{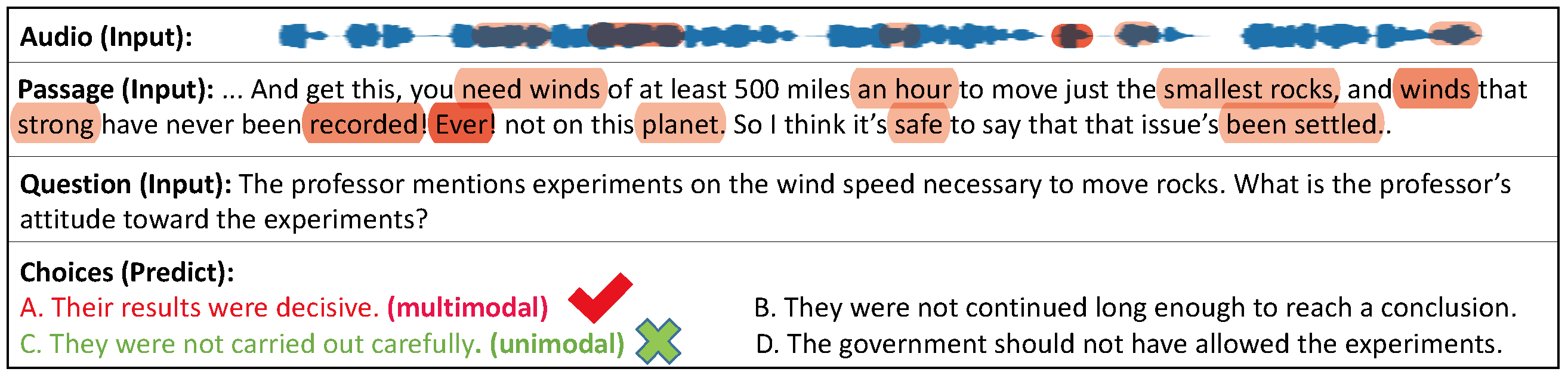}\\
		\vspace{-0.05in}
		\caption{Visualization of the attention weights on the audio and passage in the inter-modality module of our proposed Dynamic Inter- and Intra-modality Attention. 
		Each token's importance score is calculated by summing up the attention weight between this token and the tokens of other input modality. Darker color means higher weights. 
		}
		\label{fig:attention}
	    \vspace{-0.10in}
	\end{figure*}
	
	\subsection{Experimental Results}
	In this section, we will present our evaluation of the unimodal MC models, i.e., conventional Machine Reading Comprehension model, Machine Listening Comprehension model, as well as the models with our proposed MKD module, and the multimodal MC model with our proposed DIIA model.
	We conduct experiments on three representative baseline MC systems, i.e., GA Reader~\cite{Dhingra2017GAReader}, Co-Matching~\cite{Wang2018comatch}, and DCMN~\cite{Zhang2019Dual}.
    As can be seen in Table~\ref{tab:result}, under the unimodal setting, the models using text features outperform the models using audio features, which is due to the text is cleaner and easier to be processed by machines.
	Besides, our proposed MKD help the model distill multimodal knowledge in the unimodal scenarios, and can consequently achieve better performance than the conventional MC models, which verifies the effectiveness of our approach.
    From the last four rows of Table~\ref{tab:result}, we find that the fusion of multimodal features can significantly improve the performance of the model, which proves our motivation and effectiveness in proposing the audio-oriented MC task.
    In addition, compared to the shallow-fusion setting that simply add audio and textual features together, both the proposed Inter-modality Attention and Intra-modality Attention can achieve better performance by learning the inter- and intra-relationships between the two modalities, respectively, so as to obtain better multimodal representations. 
    We also implement two unimodal models (audio and text) via averaging with Co-Matching as the predictor. The ensemble model acquires accuracy of 44.64 and 52.74 on the L-TOEFL and CET datasets, while the proposed model achieves 52.03 and 57.83 accuracy, therefore the proposed model consistently outperforms the ensemble of unimodal models.
	
	Please note that our approach is not a replacement of existing approaches, but can be easily integrated into them to boost the performance, such as~\citet{Zhang2020SGNet,Xiaoguang2020Dual}.
	
	\section{Analysis}
	
	In this section, we analyze the effectiveness of the proposed DIIA and visualize the attention weights to show the advantage of the proposed approach in an intuitive manner. We also analyze the effectiveness of different passage types.
	
	\subsection{Effect of the DIIA}

	To explore the effectiveness of the multimodal features and the correlations between audio and textual features learned by DIIA, we visualize the attention weights in the inter-modality attention module.
	As shown in Figure~\ref{fig:attention}, we find that the unimodal MC system predicts the wrong answer choice `C' while the multimodal MC system predicts the correct answer choice `A'.
	To explore the reason for this difference, we visualize the attention weights in the inter-modality attention module, finding that by introducing the audio features and the DIIA model, the passage can put more attention on the key information including transitional, time and noun words, while the audio can extract more useful information, e.g., the tone information.
	The visualization of the attention weights on audio and textual information also verify our hypothesis and demonstrate the effectiveness of our approach. Since our model can capture the alignments and relationships between audio and textual modalities, the distribution of attention weights between audio and textual features is similar, which indicates that the textual features are properly enriched by the aligned audio features.
	\subsection{Types of Passage}	

	\begin{table}[t]
	\centering
	\footnotesize
	\resizebox{\linewidth}{!}{
	\begin{tabular}{l|l|cc|cc}
		\toprule
		Modalities & Methods & Con. & \(\Delta\) & Lec. & \(\Delta\)  \\
		\midrule
		\multirow{2}{*}{\begin{tabular}[c]{@{}l@{}} Unimodal \\ (Text) \end{tabular}}  & Conventional & 37.94 & - & 37.72 & - \\ \cmidrule{2-6}
		& \ w/ MKD & 47.63 & +9.69 & 46.37 & +8.65  \\ \midrule
		Multimodal & \ w/ DIIA & \bf 52.63 & +14.69 & \bf 52.17 & +14.45 \\
		\bottomrule
	\end{tabular}}
	\caption{Comparison between different type of passage and audio in the L-TOEFL dataset. Con. and Lec. stand for the Conversation passages and Lecture passages, respectively. \(\Delta\) denotes the improvement over the model under the conventional setting. 
	The models are explored with the Co-Matching \cite{Wang2018comatch} as the predictor.
	}
	\vspace{-0.12in}
	\label{citation-passagetype} 
	\end{table}

	L-TOEFL is composed of different types of passages, namely conversation and lecture, and they also correspond to different types of audio.
	Such difference may be exhibited in audio length, the number of characters or the overall audio emotions, etc.
    To better understand the improvement brought by audio information, we further explore the impacts of the passage and audio types on the MC system.
	Specifically, we conduct experiments for conversation and lecture passages on different model settings, i.e., the conventional unimodal model, the ``w/ MKD'' model, and the proposed multimodal model.
	We use the co-matching model~\cite{Wang2018comatch} as the predictor.
	Table~\ref{citation-passagetype} shows that our approach brings greater improvements in passages of conversation type than passages of lecture type. 
	We attribute this to the fact that the human mood and tone change more evidently in human conversation and are relatively smoother in lecture, which means the conversation audio can provide more useful information to the model and the multimodal MC system can predict the answer more precisely.

	\section{Conclusions}
	In this paper, we introduce the Audio-Oriented Multimodal Machine Comprehension task.
	To achieve this goal, we collect two datasets named L-TOEFL and CET, which consist of 1,324 and 876 audio-passage-question-choices pairs, respectively, and propose a Dynamic Inter- and Intra-modality Attention (DIIA) model, which consists of an inter-modality attention model and an intra-modality attention model.
	The DIIA model can learn the inter- and intra- relationships between the audio and the textual modalities. DIIA can work as an independent component and thus can be easily integrated into existing machine comprehension models.
	Furthermore, considering the abundance of unimodal (textual or audio) real-world application scenarios,
	we further develop a Multimodal Knowledge Distillation (MKD) module to enable our multimodal MC model to accurately predict the answers based only on either the text or the audio.
	The experimental results and the analysis demonstrates that audio input can improve the accuracy of machine comprehension models that solely relies on textual input. 
	Specifically, the multimodal MC models achieve better results than the unimodal MC models, 
	and the proposed DIIA model could fuse the audio and textual modalities effectively, thus boosts the baseline models by up to 21.08\% in terms of accuracy; 
	Furthermore, the MKD module allows our multimodal MC model (pretrain with both audio and textual input, predict with either audio or textual input) to outperform the unimodal models (train and predict with only audio or textual input) by up to 18.87\% in terms of accuracy.

	\section{Acknowledgments}
	Special acknowledgments are given to AOTO-PKUSZ
	Joint Research Center for Artificial Intelligence on Scene
	Cognition Technology Innovation for its support. 
	We thank all the anonymous reviewers for their constructive comments and suggestions.
	
    \bibliographystyle{aaai21}
	\bibliography{AAAI}

\end{document}